\documentclass[11pt,a4paper]{article}
\usepackage[hyperref]{acl2021}
\usepackage{times}
\usepackage{makecell}
\usepackage{latexsym}
\usepackage{graphicx}
\usepackage{amsmath}
\usepackage{amsfonts}
\usepackage{hhline}

\usepackage{microtype}

\aclfinalcopy 
\def\aclpaperid{2441} 

\title{Retrieval Enhanced Model for Commonsense Generation}

\author{Han Wang$^1$\thanks{$\;\;$Work done during internship at Microsoft.}$\;$, Yang Liu$^2$, Chenguang Zhu$^2$, Linjun Shou$^3$,\\
\textbf{Ming Gong}$^3$, \textbf{Yichong Xu}$^2$, \textbf{Michael Zeng}$^2$ \\
$^1$New York University\\
$^2$Microsoft Cognitive Services Research Group\\
$^3$STCA NLP Group, Microsoft, Beijing, China\\
  \texttt{hwang@nyu.edu}\\
  \texttt{\{yaliu10,chezhu,lisho,migon,yicxu,nzeng\}@microsoft.com} \\}
\date{}

\begin{document}
\maketitle
\begin{abstract}
Commonsense generation is a challenging task of generating a plausible sentence describing an everyday scenario using provided concepts.
Its requirement of reasoning over commonsense knowledge and compositional generalization ability even puzzles strong pre-trained language generation models.
We propose a novel framework using retrieval methods to enhance both the pre-training and fine-tuning for commonsense generation.
We retrieve prototype sentence candidates by concept matching and use them as auxiliary input.
For fine-tuning, we further boost its performance with a trainable sentence retriever.
We demonstrate experimentally on the large-scale CommonGen benchmark that our approach achieves new state-of-the-art results.\footnotemark
\end{abstract}
\footnotetext{The code and data are available at \url{https://github.com/HanNight/RE-T5}}

\section{Introduction}
The understanding of commonsense knowledge in human language has been acknowledged as a critical component for artificial intelligence systems.
In recent years, many new tasks and datasets are proposed to assess NLP model's ability of commonsense reasoning \cite{yu2020survey}.
SWAG~\cite{zellers-etal-2018-swag} is a task of inferring the upcoming event based on a partial description using commonsense.
CommonsenseQA~\cite{talmor-etal-2019-commonsenseqa}  is a commonsense question answering dataset built from ConceptNet.
Recently, \citet{lin-etal-2020-commongen} propose CommonGen, a new challenge for evaluating model's ability of generative commonsense reasoning.

CommonGen requires the system to construct a plausible sentence based on several concepts related to an everyday scenario.
Two examples for this task are shown in Table~\ref{tab:examples}.
The task is challenging because the system needs to organize provided concepts into the most plausible scenario, avoid violation of commonsense, and ensure the generated sentence is grammatically correct.
Existing approaches fine-tune pre-trained encoder-decoder models for description construction with concatenated concepts as input.

\begin{table}[t]
\renewcommand{\arraystretch}{1.1}
\begin{small}
\begin{tabular}{p{7.5cm}}
\hline
\textbf{Concept Set \#1:}                                                                                                                                                                                    \\
dog, frisbee, catch, throw                                                                                                                                                       \\ \hline
\textbf{Gold Target Sentences:}                                                                                                                                                                                   \\
A dog leaps to catch a thrown frisbee. \\
The dog catches the frisbee when the boy throws it.\\
A man throws away his dog 's favorite frisbee expecting him
to catch it in the air.\\\hhline{=}
\textbf{Concept Set \#2:}                                                                                                                                                                                    \\
lake, shore, canoe                                                                                                                                                       \\ \hline
\textbf{Gold Target Sentences:}                                                                                                                                                                                   \\
Canoe on a shore of lake.\\
Canoe on shore with rainbow across the lake.\\
Several canoes parked in the grass on the shore of a lake.\\\hline

\end{tabular}
\end{small}
\caption{Two concept sets and their gold corresponding sentences from CommonGen dataset. }\label{tab:examples}
\end{table}
 
\citet{fan-etal-2020-enhanced} propose a retrieve-and-generation method for commonsense generation which uses a prototype candidate sentence as auxiliary input.
However, their retriever is non-trainable and only works for the fine-tuning process.
In this work, we extend this idea and propose a novel framework for commonsense generation by using retrieval method for enhancing both the pre-training and fine-tuning stages.
Furthermore, we design a trainable prototype sentence retriever to further boost generation performance.

We conduct experiments on CommonGen~\cite{lin-etal-2020-commongen} benchmark dataset. 
It contains 35,141 concept sets and 79,051 corresponding sentences.
Each concept set is mapped to multiple corresponding sentences.
Without any model modification or complex fusion of knowledge graphs, our approach achieves new state-of-the-art results on CommonGen on several metrics, including BLEU, CIDEr and SPICE.

\section{Method}

We frame CommonGen challenge as a sequence-to-sequence task and adopt T5~\citep{JMLR:v21:20-074}, a powerful pre-trained encoder-decoder model, as our base model.
\citet{fan-etal-2020-enhanced} find concepts-related sentences in external corpora can benefit relational reasoning for CommonGen.
We extend this idea by proposing retrieval-enhanced T5 (RE-T5) which equips original T5 with a trainable retriever for selecting prototype sentences based on given concepts.
Meanwhile, referring to \citep{zhou2021pretraining}, we design a pre-training task for CommonGen which continue to pre-train RE-T5 on pseudo concept sets extracted from external corpora.
We also use a retriever in this pre-training stage.

Formally, given a concept set $X = \{x_1, x_2, \dots, x_n\}$, where $x_i$ represents the $i$-th concept and $n$ is the number of concepts, our goal is to generate a natural language output of tokens $Y = \{y_1, y_2, \dots, y_m\}$, which describes a common scenario in our daily life, using all given concepts in $X$. 

\begin{figure}[t]
    \begin{center}
    \includegraphics[width=0.45\textwidth]{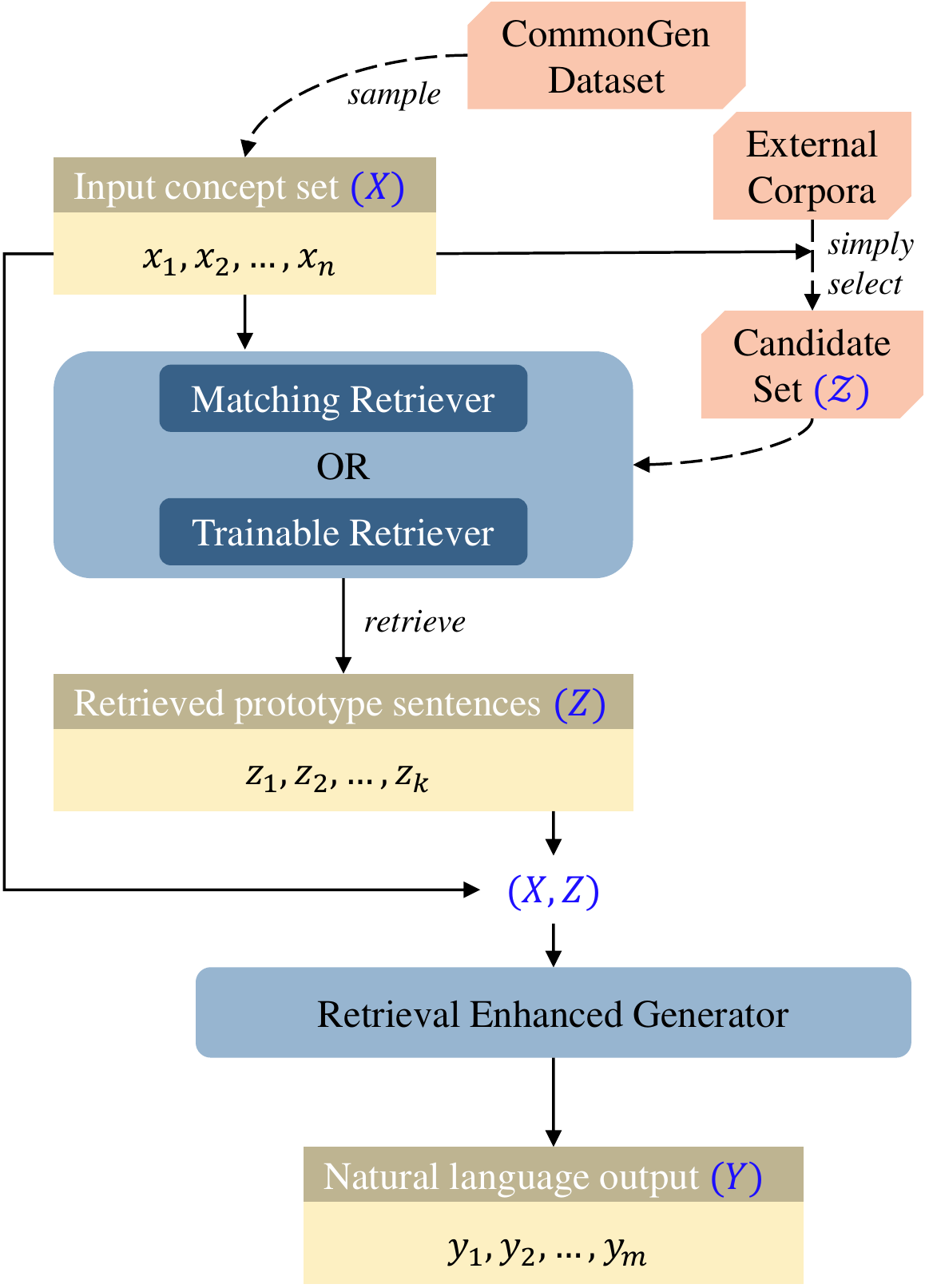}
    \caption{The overall framework of Retrieved Enhanced Model for Commonsense Generation.}
    \end{center}
    \label{fig1}
\end{figure}


\subsection{Retrieval}
Since external corpora have lots of scenario knowledge to describe the relationship between concepts~\cite{fan-etal-2020-enhanced}, we retrieve sentences related to input concepts to help the model perform better commonsense reasoning.
First, given an input concept set, we extract all sentences from external corpora that  contain at least two concepts in the input $X$ as candidate set $\mathcal{Z}$.
Then, we design two retrieval models, \textit{matching retriever} and \textit{trainable retriever}, to further retrieve $k$ prototype sentences $Z = \{z_1, z_2, \dots, z_k\}, Z\subseteq\mathcal{Z}$  as auxiliary input context for RE-T5.

\begin{table*}
\centering
\begin{tabular}{lcccc}
\hline
Model       & BLEU-4 & CIDEr  & \textbf{SPICE} & SPICE(v1.0) \\ \hline
GPT-2~\citep{radford2019language}       & 26.833 & 12.187 & 23.567         & 25.90\\
BERT-Gen~\citep{pmlr-v119-bao20a}    & 23.468 & 12.606 & 24.822         & 27.30\\
UniLM~\citep{NEURIPS2019_c20bb2d9}       & 30.616 & 14.889 & 27.429         & 30.20\\
BART~\citep{lewis-etal-2020-bart}        & 31.827 & 13.976 & 27.995         & 30.60\\
T5-base~\cite{JMLR:v21:20-074}     & 18.546 & 9.399  & 19.871         & 22.00\\
T5-large~\citep{JMLR:v21:20-074}    & 31.962 & 15.128 & 28.855         & 31.60\\
EKI-BART~\citep{fan-etal-2020-enhanced}    & 35.945 & 16.999 & 29.583         & 32.40\\
KG-BART~\citep{liu2021kgbart}     & 33.867 & 16.927 & 29.634         & 32.70\\ 
CALM(T5-base)~\citep{zhou2021pretraining}        & -      & -      & -              & 33.00\\ \hline
RE-T5 (ours)  & \textbf{40.863} & \textbf{17.663} & \textbf{31.079} & \textbf{34.30}          \\ \hline
\end{tabular}
\caption{\label{results2}
Test results on CommonGen benchmark. All results except CALM are based on the latest human references(v1.1). v1.0 indicates evaluation with old evaluation protocol.\footnotemark}
\end{table*}

\paragraph{Matching Retriever}
The \textit{matching retriever} first orders candidate sentences by the number of contained concepts.
Then it simply samples $k$ sentences starting from sentences that contained the most concepts as the auxiliary input.


\paragraph{Trainable Retriever}
In order to retrieve more useful sentences from the sentence candidate set, we design a \textit{trainable retriever}, which predicts scores to rank these candidates, and then select top-$k$ sentences as additional context. 
The scorer is built based on BERT~\citep{devlin-etal-2019-bert}, a pre-trained language model usually used for language understanding.
Given a concept set $X$ and a candidate sentence $z_i$, our trainable retriever first concatenate them into a text input:
\vspace{-0.5em}
\begin{align*}
    \texttt{[CLS]} X \texttt{[SEP]} z_i \texttt{[SEP]}\vspace{-0.5em}
\end{align*}
where  \texttt{[CLS]} and  \texttt{[SEP]} are special symbols in BERT.

We pass this into BERT, which generates an output vector for each input token.
We take the output vector  corresponding to \texttt{[CLS]} which is used as the aggregated representation of the input sequence~(denoted $\boldsymbol{c}$ ) into a linear layer with sigmoid activation to obtain the binary classification output $y_c$.
\vspace{-1em}
\begin{equation}
     y_c = \sigma(\boldsymbol{W}_c\boldsymbol{c}+b_c)\vspace{-0.5em}
\end{equation}
where $\boldsymbol{W}_c$ is a projection matrix and $b_c$ is a bias.

To train this retriever, for each concept set in CommonGen training set, we use its paired sentence as a positive example and we randomly sample another sentence, also from the training set, as a negative example.
Then, we adopt cross entropy loss for this binary classification.
The top-$k$ scored sentences with the highest scores will be selected as the auxiliary input $Z$.

We will describe how these two retrievers are used in CommonGen pre-training and fine-tuning stages.

\subsection{Pre-training}
To enhance model's ability of commonsense reasoning, we design a pre-training task for RE-T5 which is similar to original CommonGen  task.
In more details, given a sentence from external corpora, we first use
spaCy~\cite{spacy} to tag the sentences with part-of-speech and extract \textit{Verb}, \textit{Noun} and \textit{Proper Nouns} as pseudo concept phrases.
We then only keep phrases in ConceptNet~\citep{10.5555/3298023.3298212} and remove concept-sets that appear in CommonGen's testset.
We use the original sentence as the target sentence, and constructs a pre-training task of using RE-T5 to generate this sentence given pseudo concepts.

Due to the extraction method for pseudo concepts, when retrieving prototype sentences, for each concept set in pre-training data, we have a large candidate set $\mathcal{Z}$ with an excessive number of candidate sentences.
This leads to a long inference time for using the trainable retriever.
Thus, due to speed consideration and also to introduce a degree of randomness into pre-training, we use the \textit{matching retriever} to retrieve $k$ sentences as auxiliary input $Z$.

After retrieval, RE-T5 takes the concatenation of input concepts and retrieved prototype sentences as input, and the original sentence as output.


\subsection{Fine-tuning}
At fine-tuning stage, we use \textit{trainable retriever} to score sentences from candidate set $\mathcal{Z}$ and select top $k$ sentence  as additional context $Z$.
Similar to pre-training, RE-T5 takes the concatenation of input concepts and retrieved prototype sentences as input, and the original sentence as output.
\footnotetext{\url{https://inklab.usc.edu/CommonGen/leaderboard.html}}

\begin{table*}
\small
\centering
\begin{tabular}{l}
\hline
\begin{tabular}[c]{@{}l@{}}\textbf{Concept Set}:\\ trailer shirt side sit road\end{tabular}                                                                                                                                                                                                                                                                                                                                                                                                                                                                                                \\ \hline
\begin{tabular}[c]{@{}l@{}}\textbf{T5}:\\ A man sits on the side of a trailer and a shirt.\end{tabular}                                                                                                                                                                                                                                                                                                                                                                                                                                                                                    \\ \hline
\begin{tabular}[c]{@{}l@{}}\textbf{Matching Retriever}:\\ (1)Two guys in red shirts are sitting on chairs, by the side of the road, behind that open trailer.\\ (2)Two men, one wearing a straw cone hat, blue shirt, talking with a guy in a tan sunhat, red \\ plaid shirt, both with baskets in front of them, sitting on the side of a dirt road.\\ (3)An older guy with a tan shirt and hat sitting on the side of a road with bricks all around him \\ and a small green bowl on the side.\\ \textbf{RE-T5(matching retriever)}:\\ a man in a tan shirt sits on the side of a road.\end{tabular} \\ \hline
\begin{tabular}[c]{@{}l@{}}\textbf{Trainable Retriever}:\\ (1)Two guys in red shirts are sitting on chairs, by the side of the road, behind that open trailer. \\ (2)Teenagers in matching shirts stand at the side of the road holding trash bags. \\ (3)A man in a white shirt and black pants standing at the side or the road.\\ \textbf{RE-T5(trainable retriever)}:\\ a man in a white shirt and black pants sits on the side of a trailer on the road.\end{tabular}                                                                                                                       \\ \hline
\end{tabular}
\caption{\label{examples1}
An example of sentences retrieved by different retrievers and sentences generated based on them.
}
\end{table*}

\vspace{-0.5em}

\section{Experiments}
\vspace{-0.5em}

\subsection{Experiments Settings}
\paragraph{Dataset}
CommonGen is a benchmark dataset designed to diagnose whether a model has the ability of generative commonsense reasoning~\citep{lin-etal-2020-commongen}. This dataset contains 32,651/993/1,497 concept sets for training/development/test, and the numbers of  corresponding sentences are 67,389/4,018/7,644. We use BLEU~\citep{papineni-etal-2002-bleu}, ROUGE~\citep{lin-2004-rouge}, METEOR~\citep{banerjee-lavie-2005-meteor}, CIDEr~\citep{Vedantam_2015_CVPR} and SPICE~\citep{10.1007/978-3-319-46454-1_24} as evaluation metrics. Because SPICE correlates the most with human evaluation~\citep{lin-etal-2020-commongen}, we take SPICE as the primary metric.

\paragraph{External Corpora}
To be consistent with the distribution of the CommonGen dataset, we use VATEX~\citep{Wang_2019_ICCV}, Activity~\citep{Krishna_2017_ICCV}, SNLI~\citep{bowman-etal-2015-large} and MNLI~\citep{williams-etal-2018-broad} as external corpora.
We sample  500k sentences from these corpora to construct our pre-training dataset.
Meanwhile, these datasets are also used as our sentence pool for the retrieval module.
For both the pre-training and fine-tuning, all sentences that appear in the CommonGen targets are not used as retrieval sentences candidates.


\paragraph{Baselines}
We compare RE-T5 with several baseline systems.
GPT-2, BERT-Gen, UniLM, BART, and T5 are pre-trained language models tested in ~\cite{lin-etal-2020-commongen}.
They are all fine-tuned on CommonGen training set with concatenated concepts as input and description sentence as output.
EKI-BART~\cite{fan-etal-2020-enhanced} is a retrieve-and-generate framework for CommonGen, where they use a simple retriever to enhance pre-trained BART~\cite{lewis-etal-2020-bart}.
KG-BART~\cite{liu2021kgbart}  augment BART with Knowledge Graph on both the encoder and decoder side and continue to pre-train BART with a masked concept token generation task.
CALM~\cite{zhou2021pretraining} designs several self-supervised strategies encouraging model to focus on concept-centric information.

\vspace{-0.5em}

\paragraph{Implementation Details}
We adopt the T5-base as the generation model and BERT-base as the trainable retriever in fine-tuning.
We use the Huggingface Transformer~\citep{wolf-etal-2020-transformers} for model implementation. 
For pre-training phase, we use the AdamW optimizer~\citep{loshchilov2018decoupled} with an initial learning rate of 2e-6, weight decay 0.01, adam epsilon 1e-6, and a warmup fraction of 0.01. The model is pre-trained for 3 epochs, with batch size of 16, and gradient accumulation of 4 batches. For fine-tuning, the models are optimized using AdamW with an initial learning rate of 5e-5, batch size 64, gradient accumulation 3 and warmup fraction 0.01, and trained for 20 epochs.
Meanwhile, the BERT-base scorer is optimized using AdamW optimizer with an initial learning rate 2e-5, batch size 64, and the model is trained for 3 epochs. 
For the number of the retrieved sentences $k$, we experimentally choose $3$.
All experiments are conducted using 4 V100 with 32 GB memory.

\vspace{-0.5em}

\subsection{Results}

\begin{table}
\centering
\begin{tabular}{lc}
\hline
Model                      & \textbf{SPICE} \\ \hline
Retrieve (only)             & 29.60          \\ 
T5                         & 30.80\footnotemark          \\
T5 + \textit{MR}                    & 33.60          \\
T5 + \textit{MR} + pretrain         & 33.90          \\
RE-T5 (T5 + \textit{TR} + pretrain) & \textbf{34.30} \\ \hline
\end{tabular}
\caption{\label{results1}
Ablation results on the test set of CommonGen with T5-base as a backbone model. Note that \textit{MR} denotes \textit{Matching Retriever} and \textit{TR} denotes \textit{Trainable Retriever}.
}
\end{table}
Table~\ref{results2} shows results of different approaches on the CommonGen testset.
RE-T5 outperforms all previous approaches by a large margin in all metrics and sets a new state of the art.
RE-T5 combines the generation flexibility of pre-trained language models with the interpretability and modularity of a  retrieval-based approach.
Unlike EKI-BART~\citep{fan-etal-2020-enhanced} and KG-BART~\citep{liu2021kgbart}, RE-T5 enjoys strong results without model architecture modification.
It is worth noting that although T5-base baseline does not perform as well as BART~\citep{lewis-etal-2020-bart} baseline, our method still outperforms the two improved BART-based methods mentioned above.
RE-T5 demonstrates that for state-of-the-art performance, neither model modification nor complex fusion of knowledge graphs is necessary, only a simple and effective trainable retriever is needed.
\footnotetext{This is our reproduced result of T5-base. The difference from the result on the leaderboard  is also observed in other papers~\citep{zhou2021pretraining,fan-etal-2020-enhanced}.}

\vspace{-0.5em}
\paragraph{Ablation Study}
We conduct ablation experiments as shown in Table~\ref{results1}.
First, we can see that RE-T5 model outperforms the backbone T5 model by a large margin in all metrics, with 3.5 improvement in the main metric SPICE.
The second line of Table~\ref{results1} shows that, although large-scale pre-trained language models have been shown to learn and store a substantial amount of the world knowledge implicitly from the massive text corpora~\citep{petroni-etal-2019-language}, the retrieved sentences from external corpora can still explicitly expose lots of scenario knowledge to describe the relationship between concepts.
The third line indicates that further pre-training with data augmentation is helpful to improve the performance of the model.
In addition, the last line demonstrates that a trainable scorer can capture more helpful knowledge for the model for commonsense generation.

\paragraph{Example Analysis}
Through the example in Table~\ref{examples1}, we can observe that the baseline model T5 generates a sentence without concept \textit{"road"}, and the juxtaposition between \textit{"trailer"} and \textit{"shirt"} in this sentence is not in line with common sense.
For both \textit{matching retriever} and \textit{trainable retriever}, the retrieved sentences remind the model not to forget the concept \textit{"road"}, in addition to providing the relationship between \textit{shirt} and \textit{person}.
Since \textit{matching retriever} randomly retrieves sentences based on the number of concepts they contain, it tends to retrieve longer sentences to contain as many concepts as possible, which may confuse the model and thus ignore some concepts, for example, the sentence generated by RE-T5 (matching retriever) in this example is missing the concept \textit{"trailer"}.
RE-T5 (trainable retriever) can solve the above problems and generate a sentence that is fluent and in  line with common sense.

\vspace{-0.3em}

\section{Conclusions}
In this paper, we empirically investigated RE-T5, which utilizes a trainable retriever to retrieve sentences from external corpora to enhance the generative commonsense reasoning capability of pre-trained language models, such as T5.
The state-of-the-art result achieved by RE-T5 on CommonGen benchmark demonstrates that a simple yet effective trainable retriever can be a useful addition to pre-trained language models for commonsense generation. For future work, we would like to explore the possibility of extending this simple and effective retrieval-based method to more tasks. In addition, we will also try training a more advanced retrieval model to further improve the performance of commonsense generation.

\bibliographystyle{acl_natbib}
\bibliography{acl2021}



\end{document}